\pdfoutput=1

\documentclass[11pt]{article}

\usepackage[]{EMNLP2023}

\usepackage{times}
\usepackage{latexsym}
\usepackage{comment}
\usepackage{tikz}
\usepackage{pgfplots}

\usepackage[T1]{fontenc}

\usepackage[utf8]{inputenc}

\usepackage{microtype}

\usepackage{inconsolata}

\usepackage{hhline}
\usepackage{enumitem}
\usepackage{multirow}
\usepackage{tabularx, booktabs}
\usepackage{xcolor}
\usepackage{etoolbox}
\usepackage{pgf}
\usepackage{highlight}
\usepackage{graphicx}
\usepackage{amsmath,amssymb}
\usepackage{xspace}
\pgfplotsset{compat=1.18}

\newcommand{\tweetsum}{\texttt{TWEETSUMM}\xspace}
\newcommand{\todsum}{\texttt{TODSum}\xspace}
\newcommand{\samsum}{\texttt{SAMSUM}\xspace}

\newcommand{\ignore}[1]{}
\newif\ifcomment
\commenttrue 
\ifcomment
\newcommand{\agcomment}[1]{\textcolor{magenta}{\bf \small [ #1 --AG]}}
\newcommand{\cgcomment}[1]{\textcolor{blue}{\bf \small [ #1 --CG]}}
\newcommand{\mdcomment}[1]{\textcolor{blue}{\bf \small [ #1 --MD]}}
\else
\newcommand{\agcomment}[1]{}
\newcommand{\cgcomment}[1]{}
\newcommand{\mdcomment}[1]{}
\fi

\title{Evaluating Robustness of Dialogue Summarization Models in the Presence of Naturally Occurring Variations}

\author{Ankita Gupta\thanks{~~Work done in an IBM internship}~$^{\heartsuit}$ \quad Chulaka Gunasekara$^{\diamondsuit}$ \quad Hui Wan$^\clubsuit$ \quad  Jatin Ganhotra $^{\diamondsuit}$ \\ 
\bf \quad Sachindra Joshi$^\diamondsuit$ \quad Marina Danilevsky$^\diamondsuit$  \\
$^\heartsuit$University of Massachusetts Amherst, $^\diamondsuit$IBM Research AI, $^\clubsuit$Google \\ \texttt{ankitagupta@cs.umass.edu, chulaka.gunasekara@ibm.com}\\ \texttt{\{jatinganhotra,mdanile\}@us.ibm.com, jsachind@in.ibm.com}}

\begin{document}
\maketitle

\begin{abstract}
Dialogue summarization task involves summarizing long conversations while preserving the most salient information. Real-life dialogues often involve naturally occurring variations (e.g., repetitions, hesitations) and existing dialogue summarization models suffer from performance drop on such conversations. In this study, we systematically investigate the impact of such variations on state-of-the-art dialogue summarization models using publicly available datasets. To simulate real-life variations, we introduce two types of perturbations: \textit{utterance-level} perturbations that modify individual utterances with errors and language variations, and \textit{dialogue-level} perturbations that add non-informative exchanges (e.g., repetitions, greetings). We conduct our analysis along three dimensions of robustness: \textit{consistency}, \textit{saliency}, and \textit{faithfulness}, which capture different aspects of the summarization model's performance. We find that both fine-tuned and instruction-tuned models are affected by input variations, with the latter being more susceptible, particularly to dialogue-level perturbations. We also validate our findings via human evaluation. Finally, we investigate if the robustness of fine-tuned models can be improved by training them with a fraction of perturbed data and observe that this approach is insufficient to address robustness challenges with current models and thus warrants a more thorough investigation to identify better solutions. Overall, our work highlights robustness challenges in dialogue summarization and provides insights for future research.
\end{abstract}

\section{Introduction}

\begin{figure}[t]
\includegraphics[scale=0.33]{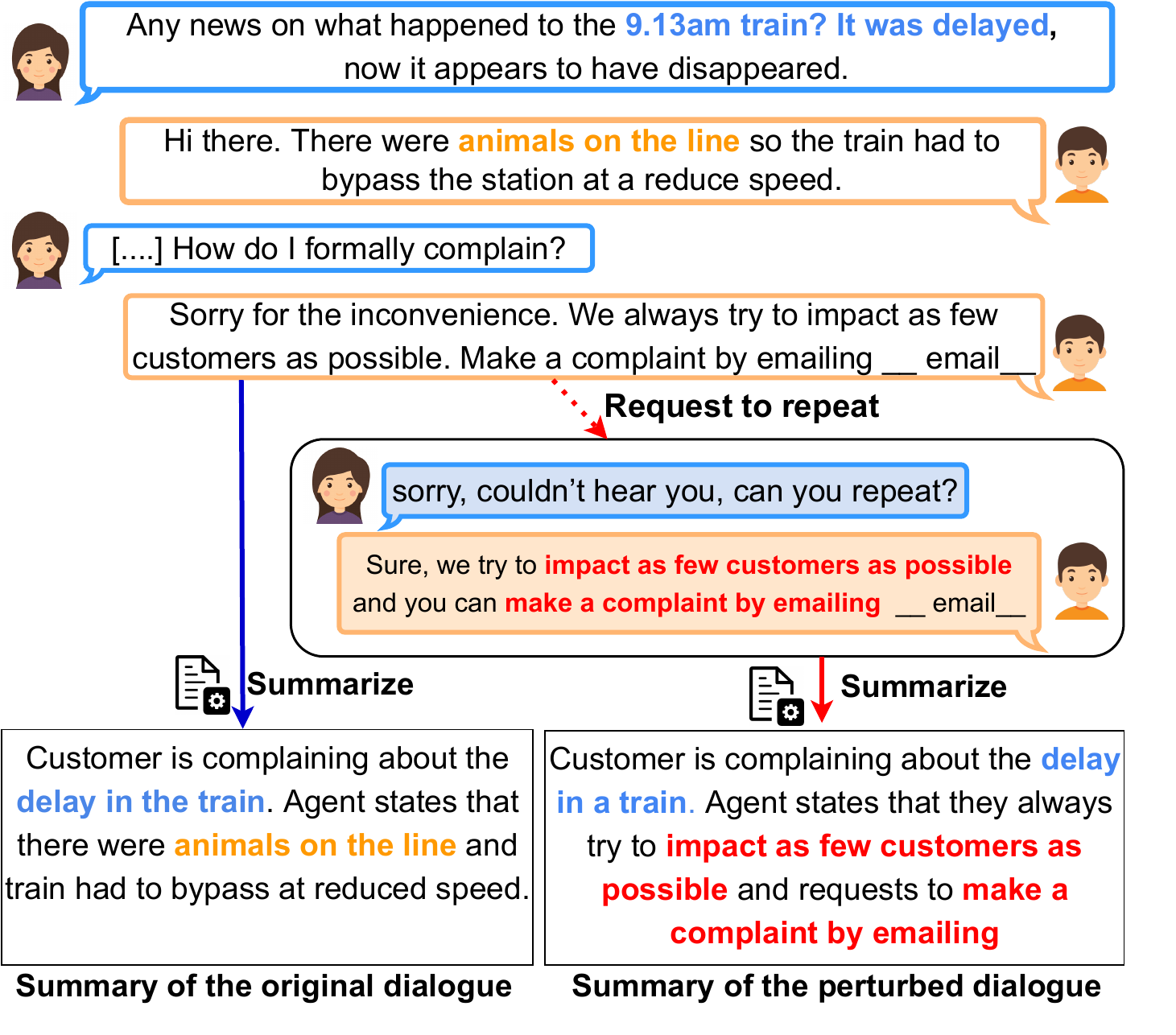}
  \caption{An example dialogue drawn from the TweetSum dataset, with a repeated utterance introduced as a perturbation. While the summary for the original dialogue includes the agent's explanation about the train delay, the summary of the perturbed dialogue includes information from the repeated utterance.}
  \label{fig:skilldetective}
  \vspace{-3mm}
\end{figure}



An increasing number of conversations are recorded and transcribed every day, spanning customer-support interactions, multi-party meetings, chit-chats among friends, etc. Deriving useful insights from such conversations requires enterprises to automatically summarize such long conversations while retaining the most salient information, also referred to as the dialogue summarization task~\cite{gliwa-etal-2019-samsum, khalman2021forumsum, feigenblat-etal-2021-tweetsumm-dialog, ijcai2022p764}. For instance, summarizing a customer-support conversation is of great value to businesses as it allows agents to write a brief summary of the conversation for record-keeping, training new agents, and decision-making purposes \cite{feigenblat-etal-2021-tweetsumm-dialog}. 

Conversations can take place in various settings, ranging from casual exchanges between acquaintances to goal-oriented business interactions involving frustrated customers and multitasking support agents. These real-life conversations often exhibit a wide range of language variations, including typographical errors, grammatical mistakes, and certain exchanges such as repetitions and speaker interruptions, which are unrelated to the primary purpose of the conversation~\cite{Sacks1974ASS}. However, existing dialogue summarization datasets, used to train current summarization models, do not adequately capture these variations, as they are typically constructed by annotators simulating specific scenarios~\cite{yuan2019abstractive} or extracted from English-speaking practice websites~\cite{gliwa-etal-2019-samsum}. Even some datasets consisting of real-life conversations~\cite{feigenblat-etal-2021-tweetsumm-dialog} might exhibit only a limited range of variations due to the nature of such conversations (e.g., Twitter conversations would lack spoken language errors). Consequently, dialogue summarization models deployed in business scenarios encounter diverse variations not observed during training. This raises a crucial question: Can current dialogue summarization models effectively handle conversations with naturally occurring variations that are legitimate inputs but not observed in the training data?

In this work, we study the impact of naturally occurring variations on the performance of the state-of-the-art dialogue summarization models using three publicly available dialogue summarization datasets. We examine the performance of a) encoder-decoder Transformer models \cite{lewis-etal-2020-bart, zhang2019pegasus, raffel-2020-exploring}, which are fine-tuned on specific dialogue summarization datasets and b) instruction-tuned models which have shown impressive zero-shot performance more recently \cite{gupta-etal-2022-instructdial, chung2022scaling}. To simulate variations, we design two kinds of perturbations: (a) utterance-level perturbations, where we make modifications to a single utterance with grammatical errors, typographical mistakes, and language-use variations, and (b) dialogue-level perturbations, where we add new utterances to the original dialogue, such that these utterances introduce no new information (e.g., repetitions, greetings). Our perturbations are inspired by the Natural Conversation Framework \cite{moore2019ncf}, grounded in observational science. This framework was created after analyzing thousands of real-world conversations across various conversational use cases and provides interactive patterns that commonly occur in real-world conversations.\footnote{Some examples include patterns such as C1.0 (opening greeting agent), C4.6 (closing success check), B2.1.0 (repeat request), A2.8 (hold request).} 

We evaluate the performance of summarization models along three conceptually different dimensions---\textit{consistency}, \textit{saliency} and \textit{faithfulness}--- each corresponding to distinct robustness issues that can arise in a dialogue summarization and elaborate on their empirical relationship. We also validate our findings via a human evaluation study.


Our analysis reveals that both fine-tuned encoder-decoder Transformer-based models and instruction-tuned models are impacted by both utterance- and dialogue-level perturbations. Instruction-tuned models are impacted more than fine-tuned models and are also more susceptible to dialogue-level perturbations than utterance-level perturbations. Both families of models show a preference for information from repeated, long, and leading utterances in the dialogue. Figure~\ref{fig:skilldetective} shows one such example where the model includes repeated utterances in the summary, whereas the non-repeated original utterance wasn't included in the summary before perturbation.  
Finally, we investigate if the performance of fine-tuned models can be improved by training with perturbed data and observe that this approach does not yield consistent performance gains, and different perturbations necessitate adding varying amounts of examples in the training data to achieve performance improvements. Thus, a more thorough investigation is needed to address these robustness challenges.

\section{Related Work}
Prior work has investigated the robustness of language understanding models mainly focusing on classification tasks \cite{Moradi2021EvaluatingTR}. Some dialogue-related classification tasks have also been explored, including dialogue act prediction \cite{liu-etal-2021-robustness}, intent detection and slot tagging \cite{einolghozati2019improving, sengupta-etal-2021-robustness}, state tracking and dialogue modeling~\cite{cho-know-2022, Tian-2021-TODDATB, Zhu2020ConvLab2AO, Kim2021HowRR, Peng2020RADDLEAE}. 

Some studies have also investigated the robustness of neural language generation models, including neural machine translation \cite{niu-etal-2020-evaluating, karpukhin-etal-2019-training, vaibhav-etal-2019-improving}, question answering \cite{peskov2019mitigating}, and open domain multi-document summarization~\cite{giorgi-2022-exploring}. However, some of these studies consider perturbations that are of extreme nature (e.g., random shuffling and deletion of words) and may occur rarely in the real world. \citet{ganhotra-etal-2020-effects} investigated the impact of natural variations on response prediction tasks in goal-oriented dialogues. 

For summarization task in particular, previous studies have focused on summarizing news articles and documents~\cite{jing-etal-2003-summarization, meechan-2019-effect, krishna-2022-improving}. However, the nature of noise in a dialogue involving multiple parties differs significantly from noise in documents. While some types of noise (e.g., spelling mistakes, grammatical errors) could occur in both, the patterns such as repetitions, reconfirmations, hesitations, and speaker interruptions \cite{Sacks1974ASS, Feng2021ASO, chen-yang-2021-simple} are peculiar to dialogues, posing unique challenges for accurate and robust summarization. The focus of this work is to assess the robustness of \textit{dialogue summarization models} in the presence of \textit{naturally occurring variations}, which has been understudied in the prior literature.

\section{Simulating Naturally Occurring Variations}
\label{sec:perturbations}
To introduce naturally-occurring variations in conversations, we experiment with two kinds of simulated perturbations, utterance-level and dialogue-level. Appendix~\ref{subsec:utterance_perturbations} provides examples for each perturbation. 


\subsection{Utterance-level Perturbations}
The utterance-level perturbations modify in a single utterance and are adapted from~\citep{liu-etal-2021-robustness}.


\paragraph{Typographical Errors}
Typographical errors occur when participants try to type quickly in chat-based interactions. We use simple regex-based perturbations, such as the removal of punctuation marks, removal or addition of whitespaces, changing letter casing, and substitutions of common expansions and contractions. We introduce spelling errors following the approach of \citet{yorke} as used in \cite{mille2021automatic}, replacing random letters with other letters closely co-located on the keyboard positions. We ensure that mistakes are not introduced in a proper-noun phrase (e.g., restaurant name) to avoid changes in important information. 


\paragraph{Grammatical Errors}
We focus on two frequent grammatical errors: dropping determiners and subject-verb disagreements. 
To drop determiners, we drop all the words in a sentence with the \textsc{DET} tag. To introduce subject-verb disagreement, we identify auxiliary verbs (via \textsc{AUX} tag) and convert between plural and singular forms as appropriate, keeping the tense unchanged. 


\paragraph{Language-use Variations}
Users can vary in their choices of dialect and vocabulary. We consider three language-use perturbations: substituting adjectives with synonyms, inflectional variations, and synthetic African American Vernacular English (AAVE) dialect. For synonym substitution, we substitute  adjectives in an utterance with their WordNet \cite{miller1998wordnet} synonyms. 
To introduce inflectional variations, we follow the approach proposed in \citet{dhole2021nlaugmenter}, where we lemmatize each content word in an utterance, randomly sample a valid POS category, and re-inflect the word according to the chosen category.
To transform an utterance to synthetic AAVE dialect, we use the set of lexical and morphosyntactic transformation rules proposed by \citet{Ziems2022VALUEUD}. 



\paragraph{Spoken Language Errors}
Spoken language errors are commonly seen in transcripts of conversations. We swap words with their homophones (e.g., \texttt{their} $\rightarrow$ \texttt{there}) to mimic speech recognition errors. We also insert filler words and speech disfluencies~\cite{Laserna2014Um} at random positions in a given utterance, spanning general filler words (e.g., \texttt{uhm}, \texttt{ah}, \texttt{err}); phrases emphasizing speaker opinion/mental state (e.g., \texttt{I believe}); and phrases indicating uncertainty (e.g., \texttt{maybe}).





\subsection{Dialogue-level Perturbations}

We craft dialogue-level perturbations by introducing new utterances that contribute no additional information, to test a model's ability to focus on the overall meaning of a conversation and identify salient information.

\paragraph{Repetitions}
Repeating and rephrasing occur commonly in real-life conversations. 
In this perturbation, we inject a synthetic utterance requesting the other participant to repeat information (e.g., \texttt{`Sorry, I couldn't hear you, can you repeat?'}). 
Since humans tend to rephrase the original message slightly instead of repeating it verbatim, we also paraphrase the original utterance before including it as a response to the request for repetition. We use the paraphraser proposed by \citet{qian-etal-2019-exploring} to paraphrase utterances. This perturbation enables us to examine repetition bias; i.e., does the model consider repeated utterances to be more significant, even when they do not contain important information. 


\paragraph{Time delays}
In customer support conversations, the agents commonly ask the customers to wait while they gather some information. 
To simulate this scenario, we add a synthetic utterance requesting to wait (e.g., \texttt{`Just give me a few minutes.'}), an acknowledgment from the other participant (e.g., \texttt{`sure'}), and finally an utterance from the first participant expressing gratitude (e.g., \texttt{`Thanks for waiting.'}). 
\paragraph{Greeting and closing remarks}
It is also common to begin a conversation with a friendly greeting and end with some closing remarks. For example, in customer support conversations, we add a greeting like \texttt{`Hi! I am your customer support assistant. How may I help you today?'} and closing remarks like \texttt{`Thank you for contacting us.'}. In open domain chit-chat, we use greetings such as \texttt{`Hey there!'} or phrases which signal end of a conversation like \texttt{`Cool, talk to you later!'}. These perturbations help us investigate structural biases present in dialogue summarization models, also known to impact news summarization models~\cite{xing-2021-demoting, jungkang-2019-biassum}. For instance, the greeting perturbation help examine lead bias (preference for the first utterance) and closing remarks help examine recency bias (preference for the last utterance).



\paragraph{Split and combined utterances}
In chat-based conversations, participants can have varying preferences for either conveying information over multiple consecutive utterances or by sending one long message. To simulate both, we split or combine utterances from the original dialogue. For example, we split an utterance into consecutive utterances by segmenting it at every five words.
Conversely, we identify sequences of consecutive utterances from a single speaker in a dialogue and concatenate them.
These perturbations allow us to examine long bias (model's preference to include a long utterance over shorter utterances, even when multiple short utterances include salient information). 


\subsection{Perturbation Quality Evaluation}
\label{sec:quality_evaluation}
To validate the assumption that our perturbations do not change the meaning of the dialogue or add any new information and to ensure the quality of our perturbed test set, we conduct a human evaluation. 
We sample 20 dialogues and their summaries from each of the three datasets (\S\ref{subsec:implementation_details}) and perturb each dialogue with all of the utterance and dialogue-level perturbations, resulting in a total of \texttt{480} dialogues. 2 annotators are asked to check whether the reference summary from the original dialogue is still a valid summary for all such perturbed dialogues (see Appendix~\ref{subsec:quality_validation} for details on annotation guidelines). For cases where the two annotators disagree, we ask a third annotator to break ties. Our annotators marked 97.5$\%$ of the perturbed dialogues as being reasonably summarized by the summary of the original dialogue,
thus validating our use of these perturbations to investigate the robustness of dialogue summarization models. This human evaluation also implies that the perturbed dialogues are readable and semantically consistent. Otherwise, for a drastically altered dialogue due to perturbations, the original summary would have been marked invalid.


\section{Quantifying Robustness}
\label{sec:quantify_robustness}

For tasks involving text generation, such as dialogue summarization, measuring robustness involves determining the relationship between different pairs of natural language texts. As a result, the robustness of generative tasks is less well-defined, compared to a classification task~\cite{liu-etal-2021-robustness} and can manifest in several ways.
We consider three dimensions for measuring robustness issues that can arise in dialogue summarization.



To facilitate subsequent discussion, we introduce the following notation: Let $x$ denote the original dialogue, $y_{r}$ be the reference summary of the original dialogue, $f$ be the summarization model trained on $(x, y_{r}) \sim D$, and $f(x)$ be its prediction over $x$. Let $x'=x+\delta$ denote the perturbed dialogue and $f(x')$ be its predicted summary. 




\paragraph{\texttt{Consistency}}
A model is consistent (and hence robust) under a perturbation ($\delta$) if the two summaries, $f(x)$ and $f(x'=x+\delta)$, are \textit{semantically similar}, resulting in minimal change. We quantify the change in model-generated output as follows,




\begin{equation}
    \Delta z_{c} = \frac{|\texttt{SCORE}(f(x), f(x))-\texttt{SCORE}(f(x), f(x'))|}{\texttt{SCORE}(f(x), f(x))}
    \label{eq:consistency}
\end{equation}
further simplified as,
\begin{equation}
    \Delta z_{c} = 1-\texttt{SCORE}(f(x), f(x'))
\end{equation}

where \texttt{SCORE} is any text similarity metric (e.g., BERTScore) that assigns a value of $1$ for identical inputs and $0$ for dissimilar inputs. 
By definition, $\Delta z_{c} \in [0, 1]$. Note that consistency is sufficient but not necessary for robustness: a good summary can be expressed in diverse ways, which leads to high robustness but low consistency. 

\paragraph{\texttt{Saliency}}
Assuming that the reference summary includes the most salient information conveyed in the input dialogue, we compute the change in salient information captured by the model-generated summaries (before and after perturbation) w.r.t the reference summary as follows:



\begin{equation}
    \Delta z_{s} = \frac{|\texttt{SCORE}(y_{r}, f(x)) - \texttt{SCORE}(y_{r}, f(x'))|}{\texttt{SCORE}(y_{r}, f(x)) }
\end{equation}

where \texttt{SCORE} is any text similarity metric (e.g., BERTScore). Since $\Delta z_{s}$ measures the normalized change in similarity scores, $\Delta z_{s} \in [0, 1]$.

\paragraph{\texttt{Faithfulness}}
Faithfulness refers to the extent to which the generated summary is supported by the content of the input dialogue, thus accurately reflecting the information without introducing spurious or fabricated details, commonly termed as hallucinations. 
We compute the change in faithfulness as follows:

\begin{equation}
    \Delta z_{f} = \frac{|\texttt{SCORE}(x, f(x)) - \texttt{SCORE}(x, f(x'))|}{\texttt{SCORE}(x, f(x))}
\end{equation}

where \texttt{SCORE} is any text-based precision metric measuring the fraction of information in the summary ($f(x)$) supported by the input dialogue ($x$) (e.g., BERTScore-Precision). Since $\Delta z_{f}$ measures the normalized change in precision scores, $\Delta z_{f} \in [0, 1]$. Note that, the second term in the numerator compares $x$ with $f(x')$ since we are interested in measuring the fraction of summary information supported by the `original dialogue.' Furthermore, since our added perturbations do not add any new information to the dialogue, $x$ and $x'$ would essentially contain the same information.  


Clearly, for all three dimensions, higher the $\Delta z$, the lower the robustness of the model. We empirically evaluate model robustness along these dimensions and discuss their relationship in $\S$\ref{subsec: dimension_correlations}. 








\section{Evaluating Robustness}

\begin{table*}[th]
\small \centering
\begin{tabular}{llrrrrrr}
\hline
\multicolumn{1}{c}{\multirow{2}{*}{\textbf{Dataset}}} &
  \multicolumn{1}{c}{\multirow{2}{*}{\textbf{Model}}} &
  \multicolumn{3}{c}{\textbf{Utterance Perturbations}} &
  \multicolumn{3}{c}{\textbf{Dialogue Perturbations}} \\ 
\multicolumn{1}{c}{} &
  \multicolumn{1}{c}{} &
  \multicolumn{1}{c}{\textbf{$\Delta z_{c}\%$}} &
  \multicolumn{1}{c}{\textbf{$\Delta z_{s}\%$}} &
  \multicolumn{1}{c}{\textbf{$\Delta z_{f}\%$}} &
  \multicolumn{1}{c}{\textbf{$\Delta z_{c}\%$}} &
  \multicolumn{1}{c}{\textbf{$\Delta z_{s}\%$}} &
  \multicolumn{1}{c}{\textbf{$\Delta z_{f}\%$}} \\ \hline
\multirow{3}{*}{TweetSum} & BART    & 17.48$\pm$0.32 & 13.37$\pm$0.68 & 24.68$\pm$1.98  & 16.77$\pm$0.40 & 10.25$\pm$2.04 & 14.48$\pm$1.98  \\
                          & Pegasus & 16.73$\pm$0.42 & 17.18$\pm$1.04 & 29.51$\pm$5.20   & 16.67$\pm$0.42 & 11.33$\pm$1.97 & 21.03$\pm$5.20   \\
                          & T5      & 17.89$\pm$0.37 & 14.44$\pm$0.82 & 16.67$\pm$2.94  & 17.02$\pm$0.38 & 11.78$\pm$1.35 & 9.81$\pm$2.94   \\ \\
\multirow{3}{*}{TODSum}   & BART    & 7.26$\pm$0.24  & 3.87$\pm$0.16  & 51.71$\pm$17.09 & 5.85$\pm$0.24  & 2.70$\pm$0.42   & 19.07$\pm$15.06 \\
                          & Pegasus & 5.20$\pm$0.21   & 3.50$\pm$0.17   & 37.85$\pm$10.74 & 3.26$\pm$0.17  & 1.74$\pm$0.32  & 22.92$\pm$19.33 \\
                          & T5      & 7.19$\pm$0.26  & 3.86$\pm$0.17  & 35.25$\pm$11.46 & 5.12$\pm$0.23  & 2.11$\pm$0.34  & 28.13$\pm$29.91 \\ \\
\multirow{3}{*}{SAMSum}   & BART    & 13.06$\pm$0.36 & 6.57$\pm$0.25  & 11.39$\pm$0.73  & 22.05$\pm$0.52 & 5.11$\pm$0.65  & 6.62$\pm$1.28   \\
                          & Pegasus & 14.21$\pm$0.39 & 6.59$\pm$0.26  & 8.21$\pm$2.05   & 20.59$\pm$0.54 & 4.35$\pm$0.5   & 6.74$\pm$5.52   \\
                          & T5      & 13.58$\pm$0.36 & 6.72$\pm$0.28  & 4.08$\pm$2.77   & 21.18$\pm$0.49 & 4.5$\pm$0.48   & 4.78$\pm$2.22   \\ \hline
\end{tabular}
\caption{Robustness evaluation of fine-tuned dialogue summarization models. The higher the score, the lower the robustness; scores are obtained using BERTScore. We observe similar trends using ROUGE-L and SummaC metrics, as mentioned in the Appendix~\ref{subsec:rouge_summac}.}
\label{table:fine-tuned-models}
\end{table*}

We present the results of our main experiments across different dialogue summarization datasets and provide key observations on how various perturbations impact the model performance.


\subsection{Implementation Details}
\label{subsec:implementation_details}
\paragraph{Datasets}
We consider two task-oriented dialogues, \tweetsum \cite{feigenblat-etal-2021-tweetsumm-dialog} and \todsum \cite{Zhao2021TODSumTD}, both consisting of conversations between an agent and a customer. \todsum comprises dialogues from multiple sub-domains (restaurants, movies, etc), collected via crowdsourcing where annotators are tasked to generate dialogues based on a given scenario. In contrast, \tweetsum focuses solely on customer support conversations occurred at Twitter. We also include \samsum \cite{gliwa-etal-2019-samsum}, a corpus of chit-chat dialogues between two or more friends. 


\paragraph{Models}
We analyze the robustness of three state-of-the-art Transformer based encoder-decoder models for dialogue summarization, \texttt{Pegasus-large} (568M parameters) \cite{zhang2019pegasus}, \texttt{BART-large} (400M parameters) \cite{lewis-etal-2020-bart} and \texttt{T5-base} (220M parameters) \cite{raffel2020exploring} models.  We choose model configurations such that the number of parameters is comparable. We fine-tune each model on the train split of the respective dialogue summarization dataset. We use beam search\footnote{We did not use nucleus sampling to avoid sampling variance.} with size 5 to generate summaries of unperturbed and perturbed dialogues. We also investigate the robustness of two instruction-tuned models, \texttt{DIAL-BART0} \cite{gupta-etal-2022-instructdial} and \texttt{FLAN-T5} \cite{chung2022scaling}, used as zero-shot summarizers, without fine-tuning on the three dialogue summarization datasets considered in this work. 

\paragraph{Metrics}
We evaluate summaries using \texttt{BERTScore}~\cite{Zhang-2020-BERTScore}, which has been shown to better correlate with human judgment~\cite{fischer-etal-2022-measuring}. \texttt{BERTScore} calculates precision, recall, and F1 scores by comparing a model-generated summary to a reference summary. We use \texttt{F1} to compute \textit{consistency} and \textit{saliency}, and \texttt{precision} to compute \textit{faithfulness}. 
To validate our observed trends, we additionally evaluate summaries using ROUGE-L metric \cite{lin2004rouge}, which measures lexical overlap, and SummaC metric \cite{laban2022summac}, which measures factual consistency. Results based on the ROUGE-L and SummaC metrics are provided in the Appendix~\ref{subsec:rouge_summac}. While we choose these metrics to report our results, the three robustness dimensions can be computed using any summarization evaluation metric. For each reported result, we use a non-parametric bootstrap~\cite[ch.~8]{wasserman2004all} to infer confidence intervals. We utilize $10^{4}$ bootstrap samples of the dialogues to report 95$\%$ bootstrap confidence intervals (CI) via the normal interval method~\cite[ch.~8.3]{wasserman2004all}.


\subsection{How robust are fine-tuned models?}
We next investigate the impact of perturbations on all three fine-tuned models across all three datasets.

\paragraph{Fine-tuned dialogue summarization models are affected by both utterance and dialogue level perturbations}
Table~\ref{table:fine-tuned-models} shows the change in \textit{consistency}, \textit{saliency}, and \textit{faithfulness} owing to utterance and dialogue level perturbations on all three datasets. All three models are equally impacted by various perturbations. 
Models trained on \texttt{TweetSum} and \texttt{SAMSum} are impacted equally by both utterance-level and dialogue-level perturbations. \texttt{TODSum} is the least impacted, since this dataset contains template-based summaries 
where only entities from the dialogue are required to be filled. 
We see a major impact on faithfulness, with the highest impact on the model trained on the \texttt{TODSum} dataset. 





\paragraph{Impact of utterance perturbations}
Table~\ref{table:utterance_perturbations} shows that utterance-level perturbations have a comparable impact (shown averaged over all three models). We also observe that the models trained on \texttt{TODSum} has little change in consistency and saliency, but a significant change in faithfulness. This is expected since the \texttt{TODSum} summaries are extractive in nature, following a pre-defined template, and only require substituting entity information extracted from the dialogue. Since the template is fixed and the summaries can only change in entity information before and after perturbation and w.r.t reference summary, we see a small change in consistency and saliency. However, we observe a large change in faithfulness, as this dimension focuses on factual correctness of the summary. 



\begin{table*}[]
\centering \small
\resizebox{0.8\textwidth}{!}{
\begin{tabular}{llrrrr}
\hline
\textbf{Dimension} & \textbf{Dataset} & \textbf{Typographical} & \textbf{Grammar} & \textbf{Language Use} & \textbf{Speech Recognition} \\ \hline
\multirow{3}{*}{$\Delta z_{c}\%$}  & TweetSum & 24.65$\pm$0.54 & 23.32$\pm$0.87 & 20.43$\pm$0.69 & 16.81$\pm$0.71 \\
                              & TODSum   & 9.97$\pm$0.30  & 5.82$\pm$0.38  & 5.73$\pm$0.28  & 4.73$\pm$0.28  \\
                              & SAMSum   & 16.27$\pm$0.36 & 16.93$\pm$0.71& 17.78$\pm$0.48& 10.88$\pm$0.52\\ \\
\multirow{3}{*}{$\Delta z_{s}\%$}     & TweetSum & 16.27$\pm$1.93 & 16.93$\pm$2.7 & 17.78$\pm$1.96 & 10.88$\pm$2.45 \\
                              & TODSum   & 5.59$\pm$1.32  & 3.12$\pm$1.04  & 2.96$\pm$0.89  & 2.49$\pm$0.98  \\
                              & SAMSum   & 7.38$\pm$2.23 & 7.44$\pm$1.54  & 7.38$\pm$1.13  & 4.76$\pm$1.02  \\ \\
\multirow{3}{*}{$\Delta z_{f}\%$} & TweetSum & 28.01$\pm$6.43 & 26.13$\pm$9.42& 19.55$\pm$8.14& 20.27$\pm$8.73\\
                              & TODSum   & 36.73$\pm$6.76& 25.30$\pm$9.81& 30.31$\pm$8.82& 18.59$\pm$9.61\\
                              & SAMSum   & 11.17$\pm$1.75& 9.98$\pm$1.83& 8.97$\pm$1.57 & 6.89$\pm$1.89 \\ \hline
\end{tabular}}
\caption{Robustness to utterance perturbations. Models are equally impacted by different perturbations.}
\label{table:utterance_perturbations}
\end{table*}


\paragraph{Impact of dialogue perturbations:}
Table~\ref{table:dialogue-bias} reports the impact of dialogue-level perturbations (shown averaged over all three models) and shows significant changes for repetition, time delays, greetings, and split utterances. 
For instance, when subjected to repetition perturbation, the models tend to include repeated utterances in the summary, even if they were previously deemed unimportant, which is referred to as repetition bias (see Figure~\ref{fig:skilldetective}). Additionally, the models demonstrate a preference for the first utterance in a dialogue (lead bias), rendering them susceptible to greetings perturbation. This observation aligns with prior findings in the field of news summarization, where sentences at the beginning of an article are more likely to contain summary-worthy information. Consequently, models trained on such datasets exhibit lead bias. Similarly, in customer-support conversations, the first utterance frequently addresses the primary issue faced by the customer. Finally, the models also prefer incorporation of lengthy utterances in the summary (long bias), by being more affected by split perturbations, and less affected by short utterances being combined.

\begin{table*}[h]
\centering \small
\resizebox{1.0\textwidth}{!}{
\begin{tabular}{llrrrrrr}
\hline
\textbf{Dimension} &
  \textbf{Dataset} &
  \multicolumn{1}{c}{\textbf{Repetitions}} &
  \multicolumn{1}{c}{\textbf{Time Delays}} &
  \multicolumn{1}{c}{\textbf{Greetings}} &
  \multicolumn{1}{c}{\textbf{Closing Remarks}} &
  \multicolumn{1}{c}{\textbf{Split}} &
  \multicolumn{1}{c}{\textbf{Combine}} \\ \hline
\multirow{3}{*}{$\Delta z_{c}\%$}  & TweetSum & 18.04$\pm$0.59 & 14.15$\pm$0.85 & 20.01 $\pm$1.34 & 9.80$\pm$1.0  & 16.71$\pm$0.83 & 6.77$\pm$0.36 \\
                              & TODSum   & 5.96$\pm$0.39  & 4.31$\pm$0.4 & 6.61$\pm$0.59  & 2.02$\pm$0.4 & 4.38$\pm$0.36 & -    \\
                              & SAMSum   & 27.32$\pm$0.46 & 22.19$\pm$0.67 & 32.89$\pm$0.99& 16.29$\pm$0.89& 11.63$\pm$0.59& 7.80$\pm$0.52\\ \\
\multirow{3}{*}{$\Delta z_{s}\%$}     & TweetSum & 12.49$\pm$3.45 & 10.53$\pm$1.47 & 15.23$\pm$5.98 & 6.03$\pm$2.23  & 11.13$\pm$1.45 &    5.40$\pm$1.34 \\ 
                              & TODSum   & 3.31$\pm$0.98  & 2.20$\pm$0.67  & 3.48$\pm$0.88  & 1.10$\pm$0.66  & 2.19$\pm$1.11  & -    \\
                              & SAMSum   & 10.87$\pm$0.23 & 8.38$\pm$0.98  & 12.63$\pm$0.95 & 6.04$\pm$1.14  & 14.65$\pm$ 0.96 &   7.05$\pm$1.26   \\ \\
\multirow{3}{*}{$\Delta z_{f}\%$} & TweetSum & 19.34$\pm$5.91 & 15.81$\pm$1.2 & 18.31$\pm$9.23 & 6.99$\pm$8.28  & 15.11$\pm$7.47 &   8.65$\pm$1.42   \\
                              & TODSum   & 64.74$\pm$6.67 & 22.74$\pm$1.66& 50.98$\pm$9.51& 10.52$\pm$9.89& 23.37$\pm$8.23& -    \\
                              & SAMSum   & 17.99$\pm$8.91 & 12.76$\pm$2.44 & 21.25$\pm$0.91 & 10.28$\pm$0.95 & 16.05$\pm$5.91  &  10.21$\pm$1.91\\
                              \hline
\end{tabular}}
\caption{Robustness to dialogue perturbations. Models are most susceptible to repetitions and time delays (repetition bias), greetings (lead bias), and split utterances (long bias). \texttt{TODSum} dataset has no consecutive utterances from the same speaker, thus we do not perform combine utterance perturbation on this dataset.}
\label{table:dialogue-bias}
\end{table*}




\subsection{Effect of model size on robustness}
Table~\ref{table:model_size} shows the change in consistency for four models, \texttt{BART-base}, \texttt{BART-large}, \texttt{T5-base}, and \texttt{T5-small}, each with a different number of parameters. The models of different sizes are almost equally affected by perturbations, suggesting that robustness issues cannot be mitigated merely by scaling the model size. 

\begin{table*}[h]
\centering \small
\resizebox{0.9\textwidth}{!}{
\begin{tabular}{lrrrrrrr}
\hline
\multirow{2}{*}{\textbf{Model}} &
  \multicolumn{1}{l}{\multirow{2}{*}{\textbf{Parameters}}} &
  \multicolumn{3}{c}{\textbf{Utterance Perturbations}} &
  \multicolumn{3}{c}{\textbf{Dialogue Perturbations}} \\ 
 &
  \multicolumn{1}{l}{} &
  \textbf{$\Delta z_{c}\%$} &
  \textbf{$\Delta z_{s}\%$} &
  \textbf{$\Delta z_{f}\%$} &
  \textbf{$\Delta z_{c}\%$} &
  \textbf{$\Delta z_{s}\%$} &
  \textbf{$\Delta z_{f}\%$} \\
  \hline
BART-large & 440 & 17.48 $\pm$0.33 & 13.37$\pm$0.68 & 24.68$\pm$0.85 & 16.77$\pm$0.40 & 10.25$\pm$2.01 & 14.48$\pm$1.98 \\
BART-base  & 140 & 18.2 $\pm$0.30  & 16.42$\pm$0.58 & 25.78$\pm$0.89 & 18.2$\pm$0.30  & 13.28$\pm$1.84 & 15.6$\pm$2.29  \\
T5-base    & 220 & 17.89 $\pm$0.37 & 14.44$\pm$0.82 & 16.67$\pm$2.94 & 17.02$\pm$0.38 & 11.78$\pm$1.35 & 9.81$\pm$2.94  \\
T5-small   & 60  & 19.15 $\pm$0.32 & 14.18$\pm$0.53 & 25.31$\pm$2.16 & 19.15$\pm$0.32 & 8.03$\pm$2.72  & 18.64$\pm$5.69 \\ \hline
\end{tabular}}
\caption{Evaluating robustness of different sized fine-tuned models on the \texttt{TweetSum} dataset.}
\label{table:model_size}
\end{table*}

\subsection{How robust are instruction-tuned models when used as zero-shot summarizers?}
\texttt{DIAL-BART0} and \texttt{FLAN-T5} have both been instruction-tuned on multiple tasks, with \texttt{DIAL-BART0}, in particular, has been instruction-tuned on dialog-specific tasks. However, neither model was trained on the \texttt{TweetSum} dataset, giving us a zero-shot setting to test their ability to summarize dialogues. As depicted in Table~\ref{table: zero-shot}, both \texttt{DIAL-BART0} ($\Delta z_{c}$=30.37$\%$ for utterance and 34.30$\%$ for dialogue) and \texttt{FLAN-T5} ($\Delta z_{c}$=38.23$\%$ for utterance and 44.12$\%$ for dialogue) are much more sensitive to perturbations compared to fine-tuned models ($\Delta z_{c}$=17.36$\%$ for utterance and 16.82$\%$ for dialogue, averaged over all three models). 

In contrast to fine-tuned models, the zero-shot models are affected more by the dialogue-level perturbations ($\Delta z_{c}$=34.30$\%$ for \texttt{DIAL-BART0} and $\Delta z_{c}$=44.12$\%$ for \texttt{FLAN-T5}) than utterance-level perturbations ($\Delta z_{c}$=30.37$\%$ for \texttt{DIAL-BART0} and $\Delta z_{c}$=38.23$\%$ for \texttt{FLAN-T5}). Among utterance-level perturbations, similar to the fine-tuned models, zero-shot models are also impacted equally by all perturbations. Among dialogue-level perturbations as well, similar to the fine-tuned models, zero-shot models are most impacted by repetitions, greetings and split utterances (Appendix \ref{subsec:zero-shot}). 



\begin{table*}[h]
\centering \small
\resizebox{0.85\textwidth}{!}{
\begin{tabular}{lrrrrrr}
\hline
& \multicolumn{3}{c}{\textbf{Utterance Perturbations}} & \multicolumn{3}{c}{\textbf{Dialogue Perturbations}} \\ 
\multirow{-2}{*}{\textbf{Model}} &
  \multicolumn{1}{c}{\textbf{$\Delta z_{c}\%$}} &
  \multicolumn{1}{c}{\textbf{$\Delta z_{s}\%$}} &
  \multicolumn{1}{c}{\textbf{$\Delta z_{f}\%$}} &
  \multicolumn{1}{c}{\textbf{$\Delta z_{c}\%$}} &
  \multicolumn{1}{c}{\textbf{$\Delta z_{s}\%$}} &
  \multicolumn{1}{c}{\textbf{$\Delta z_{f}\%$}} \\ \hline
BART                     & 17.48$\pm$0.32            & 13.37$\pm$0.68           & 24.68$\pm$1.98          & 16.77$\pm$0.40           & 10.25$\pm$2.04           & 14.48$\pm$1.98          \\
Pegasus                  & 16.73$\pm$0.42            & 17.18$\pm$1.04          & 29.51$\pm$5.20           & 16.67$\pm$0.42           & 11.33$\pm$1.97           & 21.03$\pm$5.20           \\
T5                       & 17.89$\pm$0.37            & 14.44$\pm$0.82          & 16.67$\pm$2.94           & 17.02$\pm$0.38           & 11.78$\pm$1.35           & 9.81$\pm$2.94            \\
DIAL-BART0               & 30.37$\pm$0.39           & 21.80$\pm$3.54           & 37.09$\pm$2.57           & 34.30$\pm$0.44          & 26.44$\pm$8.31        & 47.13$\pm$7.51           \\
FLAN-T5                  & 38.23$\pm$0.57           & 41.36$\pm$9.1           & 46.80$\pm$14.53           & 44.12$\pm$0.71          & 39.89$\pm$9.09           & 48.23$\pm$11.44\\ \hline          
\end{tabular}}
\caption{Robustness of zero-shot summarizers on the TweetSum dataset}
\label{table: zero-shot}
\vspace{3mm}
\resizebox{1.0\textwidth}{!}{%
\begin{tabular}{lllllllllll}
\hline
 & Typos & Grammar & \begin{tabular}[c]{@{}l@{}}Language\\ Use\end{tabular} & \begin{tabular}[c]{@{}l@{}}Spoken \\ Language\end{tabular} & Repetition & \begin{tabular}[c]{@{}l@{}}Time \\ Delay\end{tabular} & Greeting & \begin{tabular}[c]{@{}l@{}}Closing\\ Remark\end{tabular} & \begin{tabular}[c]{@{}l@{}}Split\\ Utterance\end{tabular} & \begin{tabular}[c]{@{}l@{}}Combined\\ Utterance\end{tabular} \\ \hline
\texttt{TweetSum} & 31.50 & 43.53 & 35.50 & 26.47 & 33.62 & 40.40 & 29.89 & 14.57 & 29.78 & 16.75 \\
\texttt{SAMSum} & 33.50 & 33.57 & 26.93 & 26.93 & 24.45 & 44.50 & 22.44 & 8.50 & 33.60 & 18.08 \\ \hline
\end{tabular}}
\caption{$\Delta z_{c}\%$ for dialogue and utterance-level perturbations using similarity scores from human annotators.}
\label{table:human_eval}

\end{table*}






\subsection{Validity of findings with human evaluation}
\label{subsec:trend_validity}
To validate the trends observed using the automatically computed similarity metric, we conduct a human evaluation. We use the consistency dimension for human evaluation for two main reasons: (1) Correlations: While theoretically, all three dimensions aim to measure different aspects of robustness, empirical observations reveal a strong correlation among them (Table~\ref{table:correlation}). Owing to these correlations, using any of the three dimensions would suffice for human evaluation, and (2) Ease of human evaluation: Among the three dimensions, consistency is easiest to use for human evaluation since it only requires comparison of two summaries, unlike saliency and faithfulness.
We crowdsourced similarity scores via the Appen platform\footnote{\url{https://appen.com/}} and ask annotators to compare summaries of perturbed and unperturbed dialogues, ranking their similarity on a Likert scale of 1 (highly dissimilar) to 4 (identical or paraphrases). 



To collect annotations, we utilized the same set of $20$ dialogues as in \S \ref{sec:quality_evaluation} from the \texttt{TweetSum} and \texttt{SAMSum} datasets. Each dialogue was modified by introducing perturbations from eight categories (utterance- and dialogue-level), yielding $320$ examples. 
We collected 3 annotations per example, totalling $1077$ annotations; after filtering out noisy annotations, we conducted our analysis on the remaining $514$ examples (see Appendix \ref{subsec:trend_validity_details} for annotation procedure and guidelines).

We aggregate annotations using majority voting over multiple annotations per example to get similarity predictions for each example. To compute consistency scores as per equation~\ref{eq:consistency}, we map Likert scale to continuous numeric scores from 0 to 1. Finally, we compute mean consistency scores across all examples for a given dataset and perturbation.

As shown in Table~\ref{table:human_eval}, we observe similar trends, with models exhibiting repetition, long, and lead biases and that the models are affected nearly equally by all utterance perturbations. It is important to note that while the absolute values of $\Delta z_{c}$ may differ between calculations using automatic metrics and human annotations, our focus lies in the relative impact of different perturbations on the model. Repetition, greetings, and split utterance perturbations have a greater impact on the model than combined utterance and closing remarks.

\section{Are all three dimensions necessary to measure robustness?}
\label{subsec: dimension_correlations}

While theoretically, all the three dimensions discussed in \S\ref{sec:quantify_robustness} aim to measure different aspects of robustness, empirical observations reveal a strong correlation among them. Table~\ref{table:correlation} shows the Pearson correlations\footnote{We use the SciPy's implementation available at: \url{https://docs.scipy.org/doc/scipy/reference/generated/scipy.stats.pearsonr.html}} between each pair of dimensions on the \texttt{TweetSum} dataset. Similar high correlations are also observed on the \texttt{SAMSum} and \texttt{TODSum} datasets.


This observation can be conceptually explained to some extent. For instance, high saliency implies high consistency, i.e.., if the model-generated summaries are similar to the reference summary before and after perturbation, they will be similar to each other. Similarly, high saliency implies high faithfulness, i.e.., if the model-generated summary is similar to the reference summary, it will also be factually consistent with the input dialogue (assuming the good quality of the reference summary). Thus, if $\Delta z_{s}$ is low, then $\Delta z_{c}$ is low since the summaries are close to the reference summary before and after perturbation. Furthermore, $\Delta z_{s}$ is large, then $\Delta z_{c}$ will also be large as the summaries before and after perturbation undergo change w.r.t reference summary and hence deviate from one another. Thus, $\Delta z_{s}$ and $\Delta z_{c}$ are expected to be correlated. Comparing saliency and faithfulness, if $\Delta z_{s}$ is small, then $\Delta z_{f}$ is also small. However, if $\Delta z_{s}$ is large, the model could still remain faithful (hence small $\Delta z_{f}$) under perturbation, since the summaries may be factually consistent with the input dialogue before and after perturbation yet convey very different information compared to reference summary. Thus, conceptually, there is a relation in only one direction but not the other. However, empirically these two dimensions are also correlated. 

\begin{table}[t]
\centering \small
\begin{tabular}{lccc}
\hline
\multirow{2}{*}{Model} & \multicolumn{3}{c}{Pair of dimensions} \\ 
 &
  ($\Delta z_{c}$, $\Delta z_{s}$) &
  ($\Delta z_{c}$, $\Delta z_{f}$) &
  ($\Delta z_{f}$, $\Delta z_{s}$) \\ \hline
BART                   & 0.89        & 0.91        & 0.85       \\
T5                     & 0.94        & 0.93        & 0.89       \\
Pegasus                & 0.86        & 0.85        & 0.84  \\    \hline
\end{tabular}
\caption{Pearson correlations between pairs of dimensions on the \texttt{TweetSum} dataset.}
\label{table:correlation}
\end{table}

Since all three dimensions are empirically correlated, this observation holds important implications for future robustness studies. For instance, reference summaries may not be always readily accessible, especially when assessing models in a new domain. In such situations, the consistency or faithfulness dimension can serve as a measure of robustness instead of saliency. The consistency dimension is also best suited for human evaluation for robustness studies as it depends on comparing two summaries only. In contrast, the saliency requires annotations for two ``pairs" of summaries, and the faithfulness necessitates the comparison of the summary with the dialogue, making human evaluation labor-intensive and expensive.

Visualization of correlations among the three dimensions on TwwetSum datasets is provided in Figures~\ref{fig:s_f}, \ref{fig:f_c}, and \ref{fig:s_c}. A similar analysis for SAMSum and TODSum is provided in the Appendix~\ref{sec:correlation_samsum_todsum}. 

\begin{figure}[t]
\centering
\includegraphics[scale=0.3]{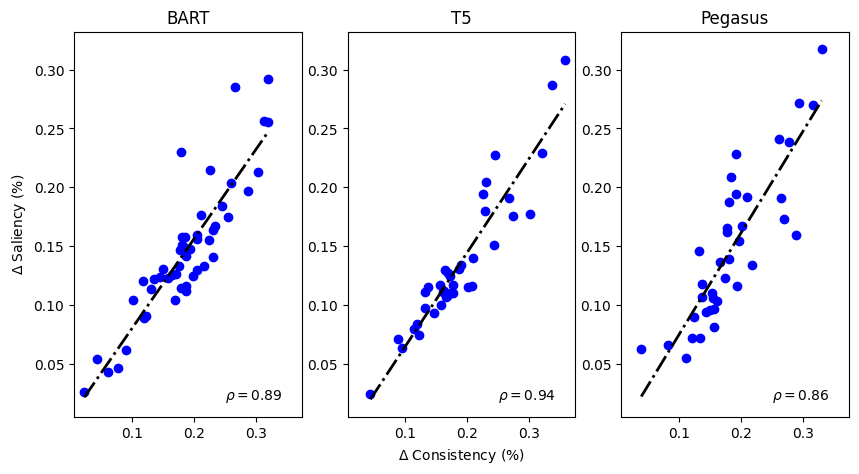}
  \caption{Correlation between consistency and saliency dimensions on TweetSum dataset.}
  \label{fig:s_c}
\end{figure}

\begin{figure}[t]
\centering
\includegraphics[scale=0.3]{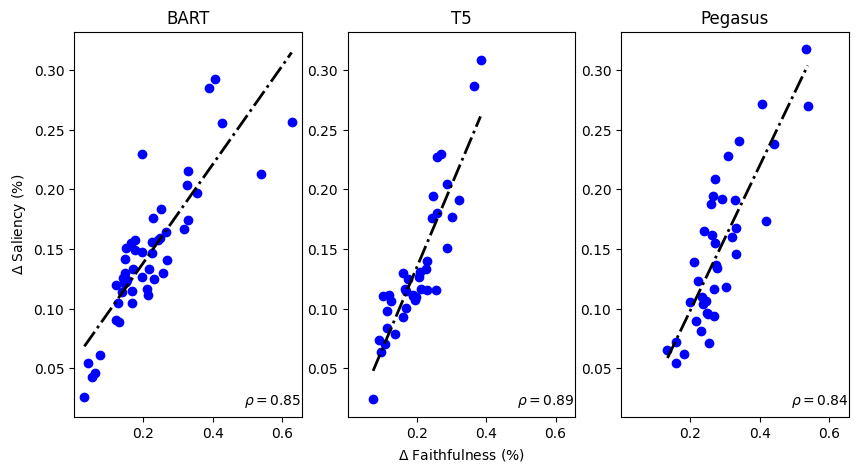}
  \caption{Correlation between faithfulness and saliency dimensions on TweetSum dataset (Outliers excluded for the purpose of visualization).}
  \label{fig:s_f}
\end{figure}

\begin{figure}[t]
\centering
\includegraphics[scale=0.3]{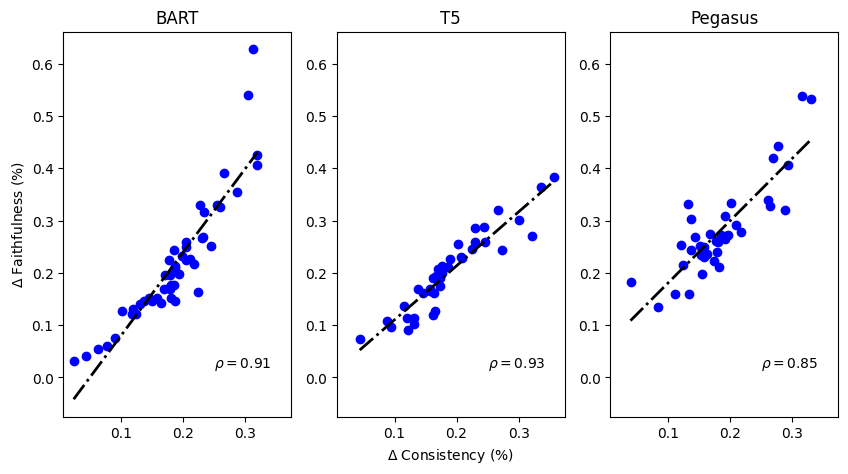}
  \caption{Correlation between faithfulness and consistency dimensions on TweetSum dataset.}
  \label{fig:f_c}
\end{figure}

\section{Improving Robustness}
To examine if training with perturbations can help mitigate robustness issues, we fine-tune \texttt{BART} on the training data augmented with perturbations and re-evaluate its performance. We create multiple training datasets, each modified by a specific kind of perturbation, using the training split of the \texttt{TweetSum} dataset. For utterance-level perturbations, we consider typographical errors and language use variations. For dialogue-level perturbations, we consider repetitions, split utterances, and greetings. For each perturbation, we modify different fractions of dialogues from the entire training set ranging from 5$\%$ to 50$\%$. We then fine-tune \texttt{BART} on all datasets and evaluate each model on the test split of \texttt{TweetSum}, which is also perturbed following the same process.\footnote{We experimented with training a single model on data with multiple perturbations and evaluating on all perturbations. However, since different perturbations can have different impacts on model performance, we found perturbation-wise analysis more interpretable.} In general, we anticipate performance improvement as we include more perturbed dialogues in the training dataset, up to a certain threshold. Beyond that point, the model would tend to overfit to the perturbations, resulting in a decline in performance.

\begin{figure}[t]
\centering
\includegraphics[scale=0.27]{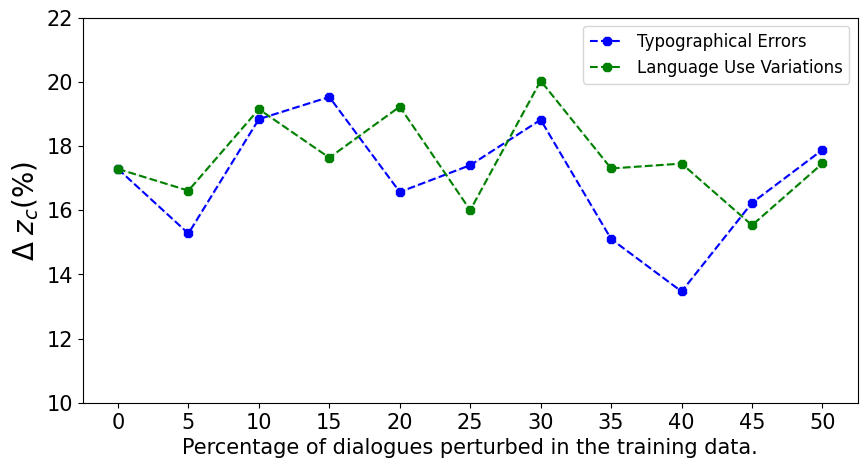}
  \caption{Fine-tuning with utterance perturbations.}
  \label{fig:fine_tuned_utterance}
\end{figure}

\begin{figure}[t]
\centering
\includegraphics[scale=0.27]{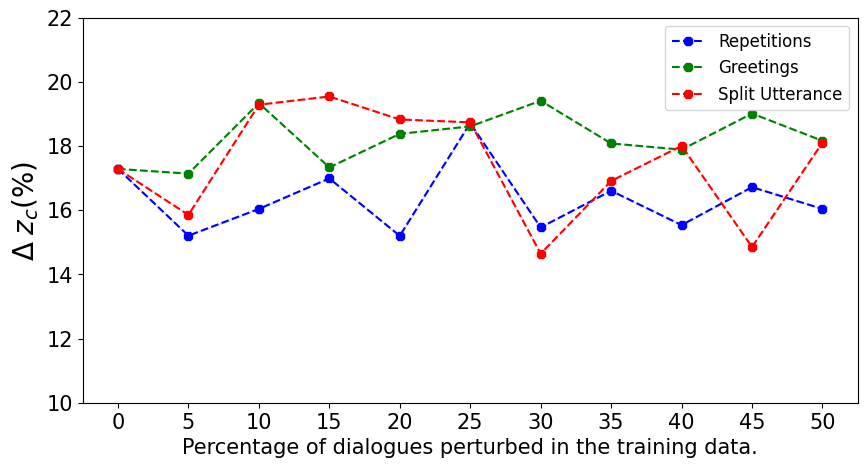}
  \caption{Fine-tuning with dialogue perturbations.}
  \label{fig:fine_tuned_dialog}
\end{figure}

Figure~\ref{fig:fine_tuned_utterance} and Figure~\ref{fig:fine_tuned_dialog} show the change in model consistency when fine-tuned with perturbations. The lower the change in consistency, the higher the model robustness to the perturbations. One takeaway is that different perturbations necessitate varying amounts of perturbed examples in the training set to achieve maximum performance improvement. For example, typographical errors and language use variations yield the largest drop in $\Delta z_{c}$ when approximately 40$\%$ and 45$\%$ of the dialogues are perturbed during training. In contrast, dialogue-level perturbations require significantly less perturbed data during training, with approximately 30$\%$ split-utterances, 15$\%$ greetings, and only 5$\%$ repetitions being sufficient. Overall, the results demonstrate that fine-tuning with perturbed data does not yield consistent performance improvements, and more detailed exploration needs to be conducted as part of future work. 

\section{Can we remove perturbations using reverse heuristics or language models to address robustness issues?}
\paragraph{Removing perturbations using reverse heuristics}
A potential solution to address the robustness issues can be to use reverse heuristics to remove certain perturbations from the dialogues (e.g., greetings). However, not all of the perturbations we consider are easily discoverable and removable. For instance, in repetitions or time-delay perturbations, the repeated utterance could include more or less information than the original utterance. As a result, using heuristics that preserve only the original or the repeated utterance can not only affect the dialogue's readability but also impact the information conveyed in the dialogue.

We acknowledge that greetings and closing remarks can be simpler to remove via reverse heuristics. However, our primary purpose in introducing greetings and closing remarks perturbations is to study model behavior (e.g., potential lead and recency biases). Introducing greetings and conclusions is just a means to an end, i.e., a systematic approach to investigate these biases, illustrating the model’s preferences towards first and last utterances in a dialogue. There can be many scenarios where a non-greeting first utterance is not salient in the dialogue compared to other utterances, yet it is included in the summary just because of its position in the dialogue. The space of such generic utterances is vast and, hence difficult to operationalize. Thus, we propose to use greetings/closing remarks, which occur naturally in real-life conversations and provide a simple way to investigate these biases.

\paragraph{Removing perturbations using language models}
Another potential solution to address robustness issues can be to use language models (LM) to preprocess dialogues to remove errors and repetitions. Such an approach could suffer from two potential challenges: (1) In a deployment setting, such a strategy could increase latency since it will require a) dialogue pre-processing via LM followed by b) dialogue summarization, instead of directly summarizing the dialogue. (2) LMs are known to hallucinate content, and such pre-processing poses the risk of introducing unwanted factual errors in the input dialogue.

\section{Conclusion}
We investigate the impact of naturally occurring variations on state-of-the-art dialogue summarization models using three publicly available datasets. To simulate variations, we introduce utterance-level and dialogue-level perturbations. We conduct our analysis using three dimensions of robustness: consistency, saliency, and faithfulness, which capture different aspects of summarization model performance. Our results show that both fine-tuned and instruction-tuned models are affected by perturbations, with instruction-tuned models being more susceptible, particularly to dialogue-level perturbations. We also observe that both model families exhibit repetition, long, and lead biases. We confirm our findings via human evaluation. Finally, we show that training with perturbed data improves the robustness of fine-tuned models.
\section{Limitations}
We list some of the limitations of our study which researchers and practitioners would hopefully benefit from when interpreting our analysis. 1) Our analysis uses automatic metrics to measure semantic similarity. Established metrics such BERTScore are imperfect \cite{deutsch-etal-2022-examining}. However, they are widely used in the summarization literature, and also correlate with human judgements of summary quality, and thus are useful for comparing system-level performance. To validate our findings, we also conduct a human evaluation to better understand trends observed due to various perturbations. The investigation of better-automated metrics for natural language generation is an active field of research, and we hope to integrate novel performance metrics in future work. (2) While our perturbations are motivated by real-life scenarios, they are still synthetic in nature. However, we take care wherever possible to avoid unrealistic changes to the dialogues. (3) Our study limits to only open-sourced models and does not investigate the robustness of proprietary LLMs (e.g., ChatGPT), which may be more robust. We decided to limit our study to open-sourced models as it allows us to carefully control what is in the training data, which is not possible with proprietary LLMs and the possibility of data contamination also makes it hard to draw conclusions. (4) Our study does not include spoken conversations, which would bring in very different and diverse nuances of spoken conversations compared to text-based conversations, and is currently out of the scope of this paper. (5) Our study proposes one possible method to measure robustness, and we acknowledge that there can be many other viable ways to quantify robustness. However, quantifying the robustness of tasks involving text generation (e.g., summarization) is an active area of research \cite{wang-etal-2022-measure} and we hope our work will spur further investigation as part of future work. (6) We did not investigate the robustness of models under both utterance and dialogue level perturbations occurring together in a single dialogue, as that would result in a large number of possible combinations to consider. We leave this for future work.





\section{Ethics Statement}
All annotators in our human evaluation were recruited via Appen platform and were presented with a consent form prior to the annotation. They were also informed that only satisfactory performance on the screening example will allow them to take part in the annotation task. None of the material/examples they looked at had any hateful or abusive content. We also ensured that the annotators were paid fair amount of wages using Appen's Fair Pay Price Per Judgment which equates to an hourly rate matching a little over the minimum wage of annotators in their respective countries.

\bibliography{custom}
\bibliographystyle{acl_natbib}

\newpage
\appendix

\section{Appendix}
\label{sec:appendix}
\subsection{Details/Examples of Perturbations}
\label{subsec:utterance_perturbations}

See Table~\ref{table:perturbations_examples}.

\begin{table*}[]
\centering \tiny
\begin{tabular}{llll}
\hline
Perturbation Type &
  Perturbation Category &
  Perturbation Name &
  Examples \\ \hline
\multirow{11}{*}{Utterance Level} &
  \multirow{5}{*}{Typographical Errors} &
  remove punctuation &
  \texttt{great!}$\rightarrow$ \texttt{great} \\
 &
   &
  remove/add whitespace &
  \texttt{Customer} $\rightarrow$ \texttt{Custo mer} \\
 &
   &
  change letter casing &
  \texttt{action} $\rightarrow$ \texttt{actIon} \\
 &
   &
  common substitutions expansions &
  \texttt{n’t} $\rightarrow$ \texttt{not} \\
 &
   &
  common substitutions contractions &
  \texttt{I am} $\rightarrow$ \texttt{I’m} \\
 &
  \multirow{2}{*}{Grammatical Errors} &
  dropping determiners &
  \texttt{a, the, an} \\
 &
   &
  subject-verb disagreements &
  \texttt{She likes apples.} $\rightarrow$ \texttt{She like apples.} \\
 &
  \multirow{4}{*}{Spoken Language Errors} &
  homophone swaps &
  \texttt{their} $\rightarrow$ \texttt{there} \\
 &
   &
  \multirow{3}{*}{filler words and disfluencies} &
  \begin{tabular}[c]{@{}l@{}}\texttt{uhm}, \texttt{uh}, \texttt{erm}, \texttt{ah}, \texttt{er}, \texttt{err}, \\ \texttt{actually}, \texttt{like}, \texttt{you} \texttt{know}\end{tabular} \\
 &
   &
   &
  \texttt{I think/believe/mean}, \texttt{I would say} \\
 &
   &
   &
  \begin{tabular}[c]{@{}l@{}}\texttt{maybe}, \texttt{perhaps}, \texttt{probably}, \texttt{possibly}, \\ \texttt{most likely}\end{tabular} \\ \hline
\multirow{8}{*}{Dialogue Level} &
  Repetitions &
  N/A &
  \texttt{`Sorry, I couldn't hear you, can you repeat?'} \\
 &
  \multirow{3}{*}{Time Delays} &
  \multirow{3}{*}{N/A} &
  \texttt{`Just give me a few minutes..'} \\
 &
   &
   &
  \texttt{`sure'}, \texttt{`yup!'} \\
 &
   &
   &
  \texttt{`Thanks for waiting.'} \\
 &
  \multirow{4}{*}{Greeting and closing remarks} &
  greeting (Customer Support) &
  \texttt{`Hi! I am your customer support assistant. How may I help you today?'} \\
 &
   &
  greeting (friends) &
  \texttt{`Hi!'} or \texttt{`Hey there!'} \\
 &
   &
  closing (Customer Support) &
  \texttt{`Thank you for contacting us. Have a nice day!'} \\
 &
   &
  closing (friends) &
  \texttt{`Cool, talk to you later!'}, \texttt{`Bye.'} \\ \hline
\end{tabular}
\caption{Examples of each perturbation}
\label{table:perturbations_examples}
\end{table*}






\subsection{Details of annotation guidelines of quality validation in \S5.2}
\label{subsec:quality_validation}
\begin{figure*}[t]
\centering
\includegraphics[scale=0.3]{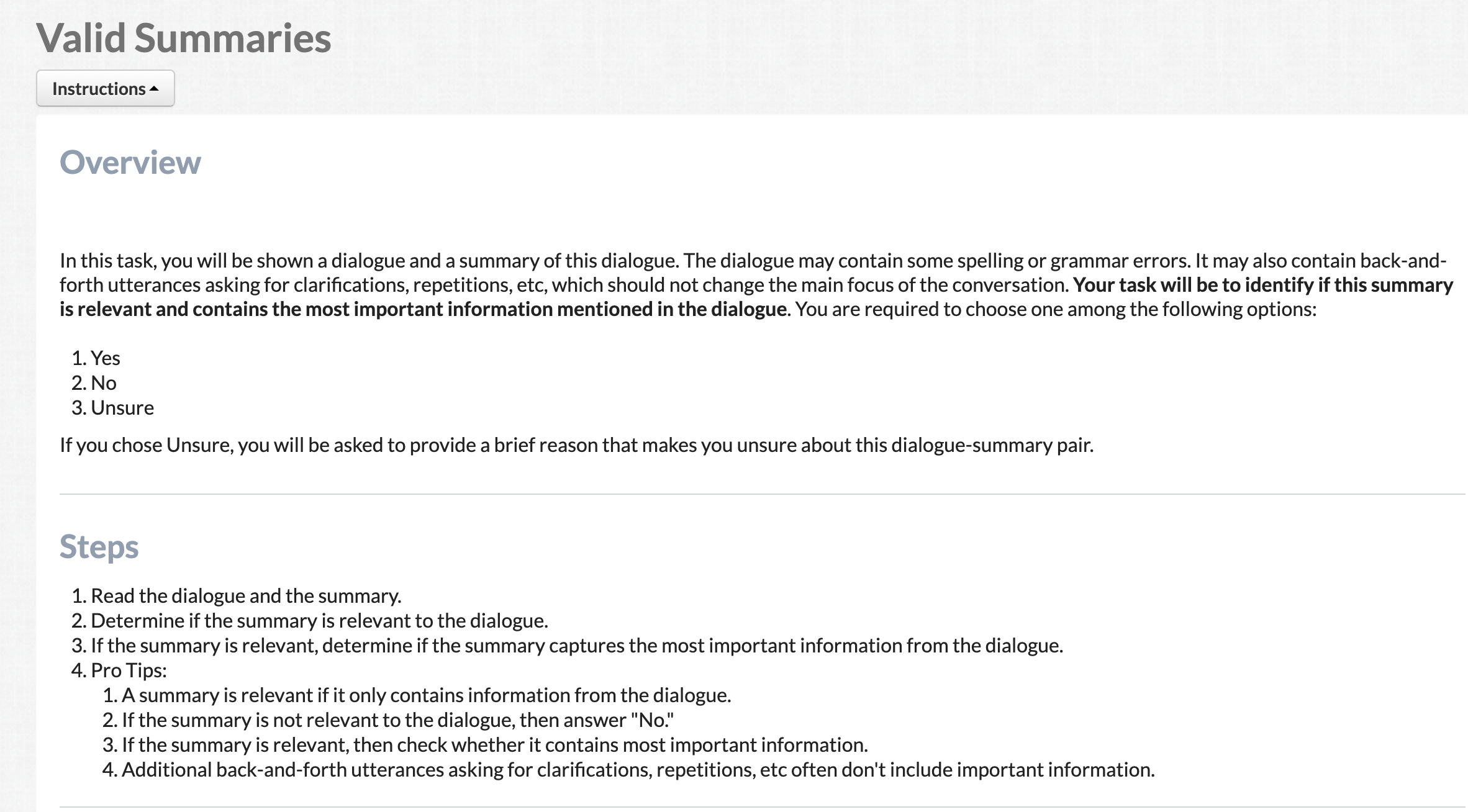}
  \caption{Annotation guidelines for quality validation of perturbed dialogue-summary pairs.}
  \label{fig:task1}
\end{figure*}

\begin{figure*}[t]
\centering
\includegraphics[scale=0.3]{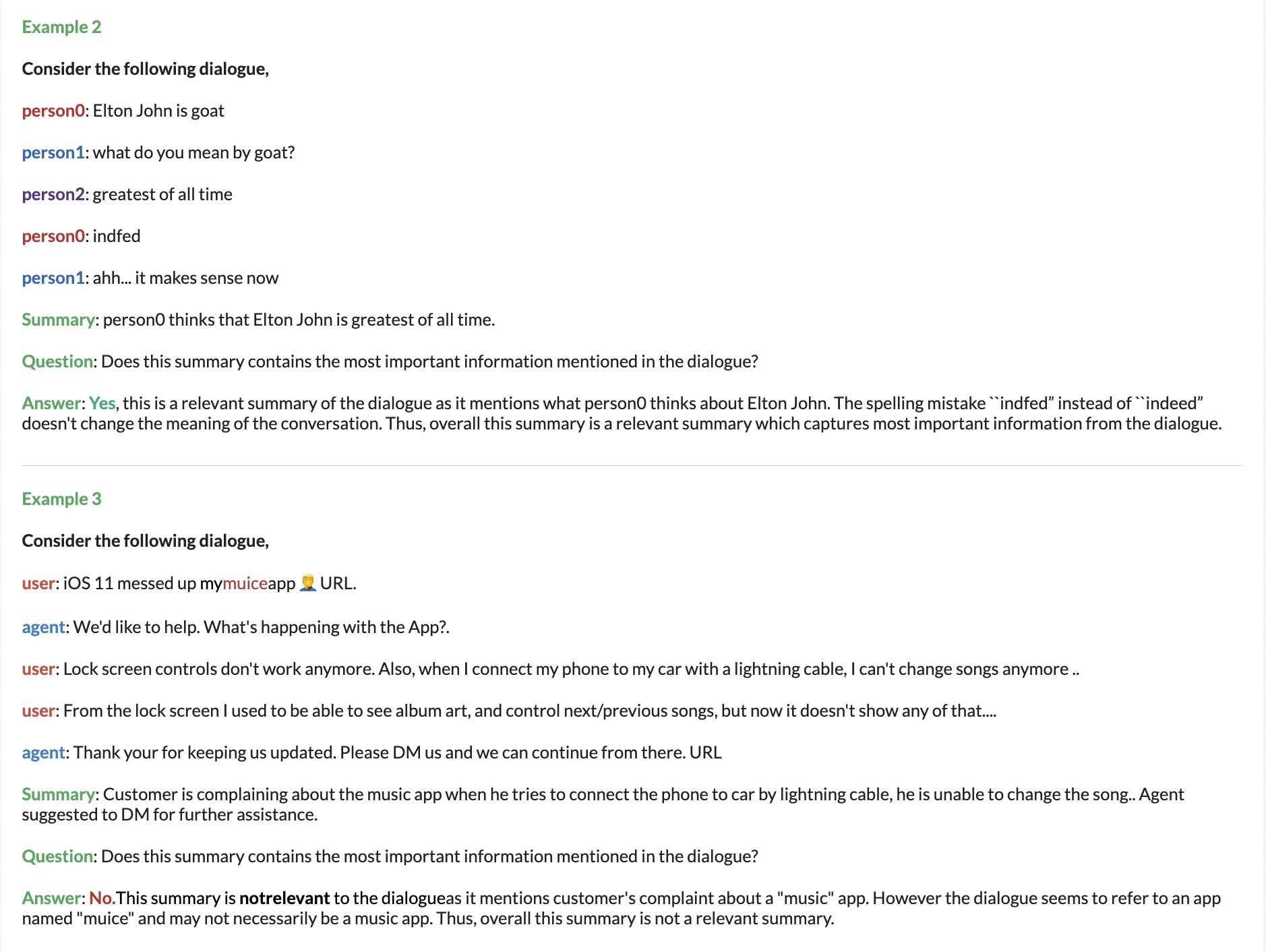}
  \caption{Examples provided as part of annotation guidelines for quality validation of perturbed dialogue-summary pairs}
  \label{fig:task1_examples}
\end{figure*}

For annotation collection, we only allowed annotators proficient in English from a small group of the most experienced annotators adjudicated by the Appen platform; from any country. We also used hidden test questions for quality control and required annotators to maintain at least $80\%$ accuracy throughout the job on these hidden test questions. These test questions are pre-labeled and are used before and during the task to quiz the annotator. We selected 15 test questions from the validation split of each dataset ensuring that these questions do not overlap with questions seen by the annotators for the actual annotation task. Figure~\ref{fig:task1} shows the annotation guidelines and Figure~\ref{fig:task1_examples} shows examples provided for this task. 

\subsection{Details of annotation guidelines for the validity of trends in \S5.6}
\label{subsec:trend_validity_details}
\begin{figure*}[t]
\centering
\includegraphics[scale=0.3]{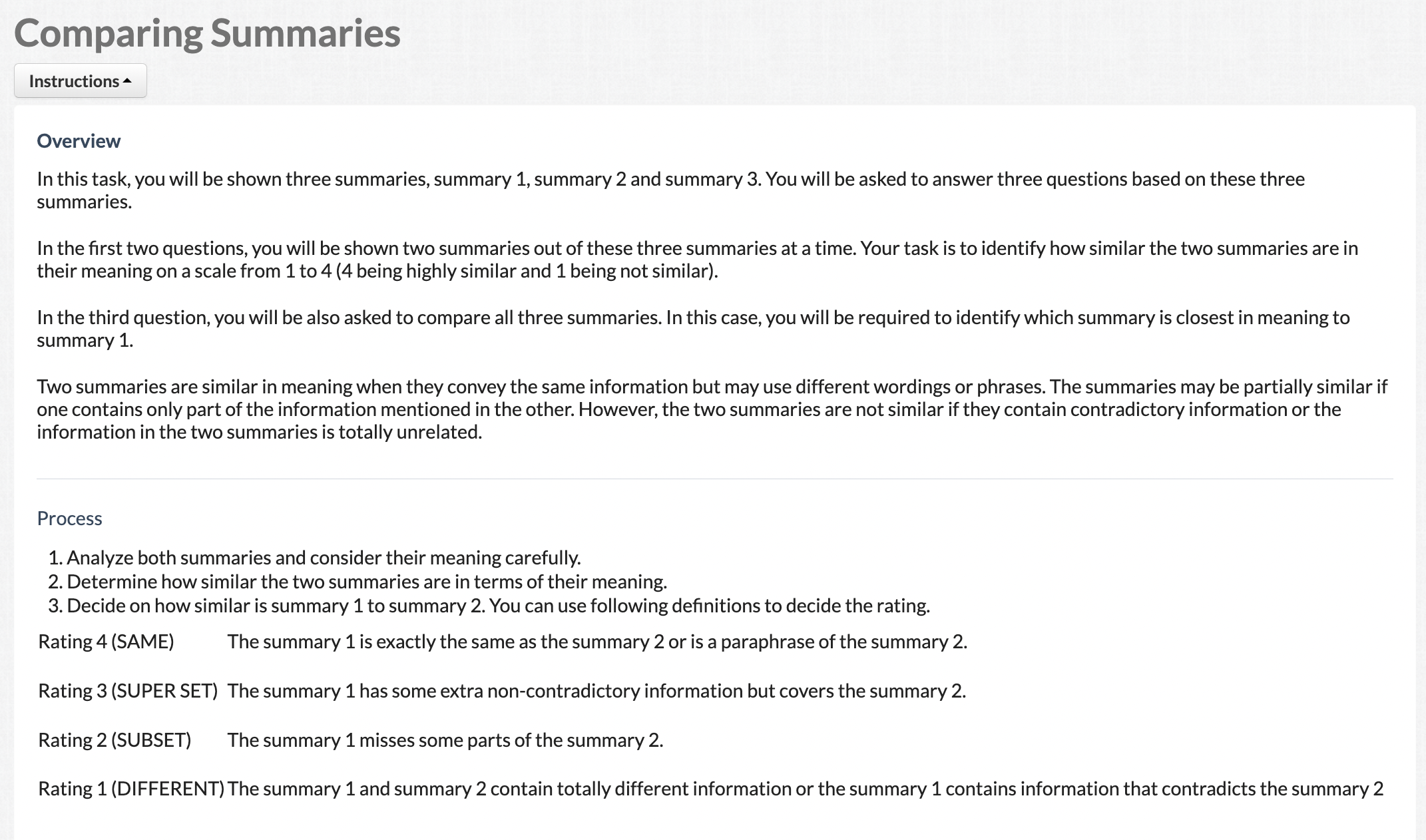}
  \caption{Annotation guidelines for the validity of trends; to collect similarity annotations for pair of summaries.}
  \label{fig:task2}
\end{figure*}

\begin{figure*}[t]
\centering
\includegraphics[scale=0.3]{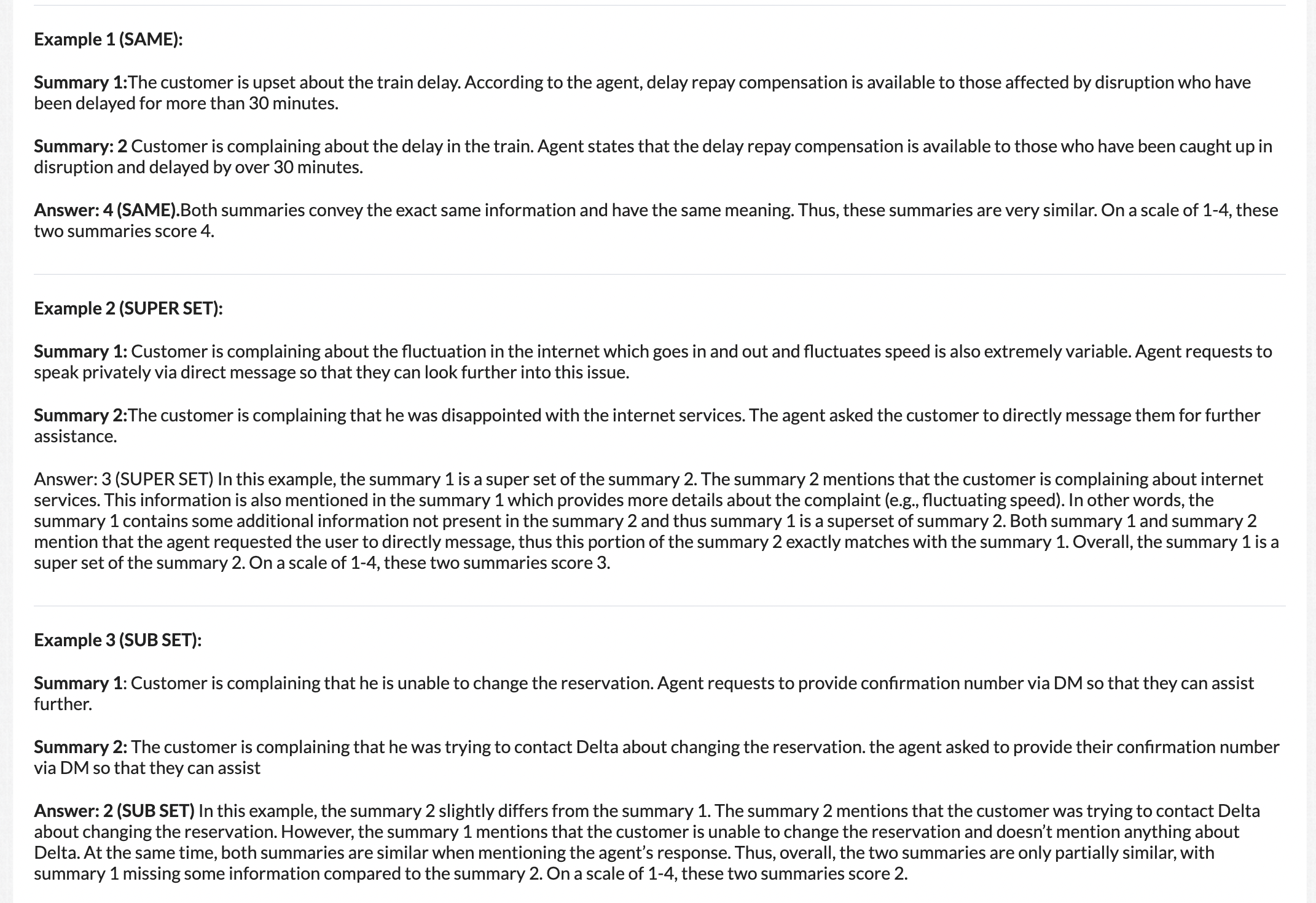}
  \caption{Examples provided as part of annotation guidelines to collect similarity annotations for pair of summaries.}
  \label{fig:task2_examples}
\end{figure*}

\paragraph{Quality Control:} For this task, as well we only allowed annotators proficient in English from a small group of the most experienced annotators adjudicated by the Appen platform; from any country. We also used hidden test questions for quality control and required annotators to maintain at least $80\%$ accuracy throughout the job on these hidden test questions. Figure~\ref{fig:task2} shows the annotation guidelines, and Figure~\ref{fig:task2_examples} shows examples provided for this task. 

\paragraph{Number of annotations:} In the main task, each annotator was shown 5 examples per page with one hidden test example. For each example, we collected three annotations. In cases where there was no agreement among the initial three annotations, we obtained additional annotations. A maximum of five annotations was considered. 

\paragraph{Noise Filtering:} Before computing consistency scores, we took several steps to filter out noisy annotations. The Appen platform estimates the trust score for each worker (by calculating accuracy on hidden test examples) and also marks examples as tainted if it is annotated by an annotator whose accuracy score has fallen below the minimum accuracy threshold. To retain only the highest quality annotations, we remove annotations that were marked as tainted and only keep annotations from workers whose trust score is 100$\%$, resulting in $795$ annotations. On qualitatively examining the annotations we also found cases where the two summaries were word-by-word the same, yet the annotator did not give a rating of $4$ (highly similar or exact match). Since this is a case of obvious noise, we remove such cases. If an example has less than 3 annotations left after the filtering step, we drop the example.
After this filtering, we finally use $514$ annotations to conduct our analysis.

\subsection{Targeted dialogue perturbations to investigate the repetition bias}
\label{subsec:targeted_perturbations}

\begin{table}[h]
\centering \tiny
\begin{tabular}{llrrr}
\hline
\multirow{2}{*}{\textbf{Dataset}} &
  \multirow{2}{*}{\textbf{Model}} &
  \multicolumn{3}{c}{\textbf{Repeated Utterance}} \\ 
 &
   &
  \textbf{Most Relevant} &
  \textbf{Least Relevant} &
  \textbf{Random} \\ \hline
\multirow{3}{*}{TweetSum} & BART    & 12.40 & 14.53 & 14.46 \\
                          & Pegasus & 13.49 & 16.68 & 14.22 \\
                          & T5      & 9.26  & 11.46 & 10.84 \\ \\
\multirow{3}{*}{TODSum}   & BART    & 1.94  & 4.32  & 3.52  \\
                          & Pegasus & 2.05  & 2.05  & 2.92  \\
                          & T5      & 1.85  & 3.66  & 3.50 \\
                          \hline
\end{tabular}
\caption{Saliency scores of fine-tuned models with targeted perturbations. Perturbing the least relevant utterance results in the highest change in saliency, suggesting that the model exhibits repetition bias. }
\label{table:targeted_perturbations}
\end{table}
To delve deeper into the repetition bias observed in the models, we conducted targeted perturbations, where we repeat utterances based on whether the information conveyed in those utterances was considered important by the reference summary. Specifically, we identify utterances that are highly relevant and least relevant to the reference summary. To measure relevance, we compute semantic similarity\footnote{using sentence transformers [CITE]} between each utterance and each sentence in the reference summary. For each summary sentence, we then determine the most (least) relevant utterance by selecting the one with the highest (lowest) similarity with the summary sentence. When perturbing the most relevant utterance, we perturb the utterances that were identified as relevant to at least one summary sentence. When perturbing the least relevant utterance, we perturb the utterances that were identified as least relevant to all the summary sentences. 

As shown in Table~\ref{table:targeted_perturbations}, we observe that the model exhibits the highest change in saliency scores when we perturb the least relevant utterance, which further demonstrates the model's tendency to consider repeated information as important, even though it was not considered important as per the reference summary. In contrast, repetition of the most relevant utterance shows the least change in the scores, since the model already focuses on the most relevant information before perturbation and after repeating that utterance, it still remains important to be included in the summary.

\subsection{Perturbation-wise impact on zero-shot models}
\label{subsec:zero-shot}
See Table~\ref{table:zero-shot-dialogue} and Table~\ref{table:zero-shot-utterance} 

\begin{table*}[]
\centering \tiny
\begin{tabular}{crrrrrr}
\hline
\multirow{2}{*}{Model}                  & \multicolumn{6}{c}{\textbf{Perturbation}}          \\ \cline{2-7} 
 &
  \textbf{repetitions} &
  \textbf{time\_delays} &
  \textbf{greetings} &
  \textbf{Closing remarks} &
  \textbf{split\_utterances} &
  \textbf{combined\_utterances} \\ \hline
\multicolumn{1}{r}{\textbf{DIAL-BART0}} & 35.30 & 31.15 & 35.02 & 23.07 & 35.10 & 18.31 \\
\multicolumn{1}{r}{\textbf{FLAN-T5}}    & 45.65 & 32.88 & 60.10 & 48.11 & 41.45 & 20.34 \\ \hline
\end{tabular}
\caption{Change in consistency scores due to dialouge-level perturbations on instruction-tuned models when used as zero-shot summarizers. Models are more affected due to repetitions, time-delays, greetings, and split utterances compared to closing remarks and combined utterances.}
\label{table:zero-shot-dialogue}
\end{table*}

\begin{table*}[]
\centering \tiny
\begin{tabular}{crrrr}
\hline
\multirow{2}{*}{Model}                  & \multicolumn{4}{c}{\textbf{Perturbation}} \\ \cline{2-5} 
 & \textbf{typographical} & \textbf{grammar} & \textbf{language\_use} & \textbf{speech\_recognition} \\ \hline
\multicolumn{1}{r}{\textbf{DIAL-BART0}} & 33.74  & 32.26 & 27.53 & 30.33 \\
\multicolumn{1}{r}{\textbf{FLAN-T5}}    & 42.60  & 48.03 & 39.75 & 33.86 \\ \hline
\end{tabular}
\caption{Change in consistency scores due to utterance-level perturbations on instruction-tuned models when used as zero-shot summarizers. Models are equally affected due to all perturbations.}
\label{table:zero-shot-utterance}
\end{table*}

\subsection{Correlation analysis}
\label{sec:correlation_samsum_todsum}
\subsubsection{SAMSUM}
\begin{figure}[t]
\centering
\includegraphics[scale=0.3]{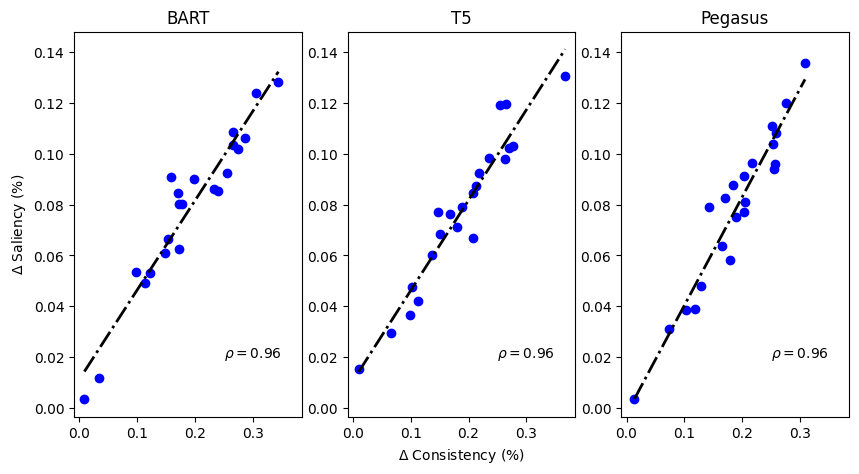}
  \caption{Correlation between consistency and saliency dimensions on SAMSum dataset.}
  \label{fig:s_c_sam}
\end{figure}

\begin{figure}[t]
\centering
\includegraphics[scale=0.3]{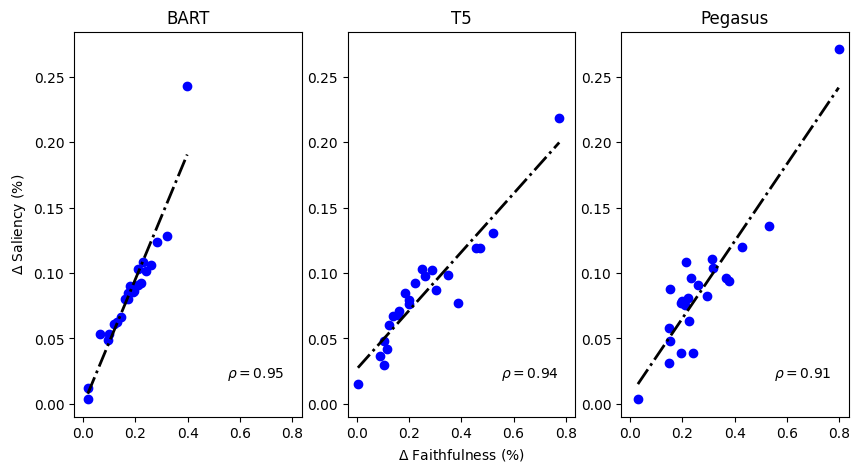}
  \caption{Correlation between faithfulness and saliency dimensions on SAMSum dataset (Outliers excluded for the purpose of visualization).}
  \label{fig:s_f_sam}
\end{figure}

\begin{figure}[t]
\centering
\includegraphics[scale=0.3]{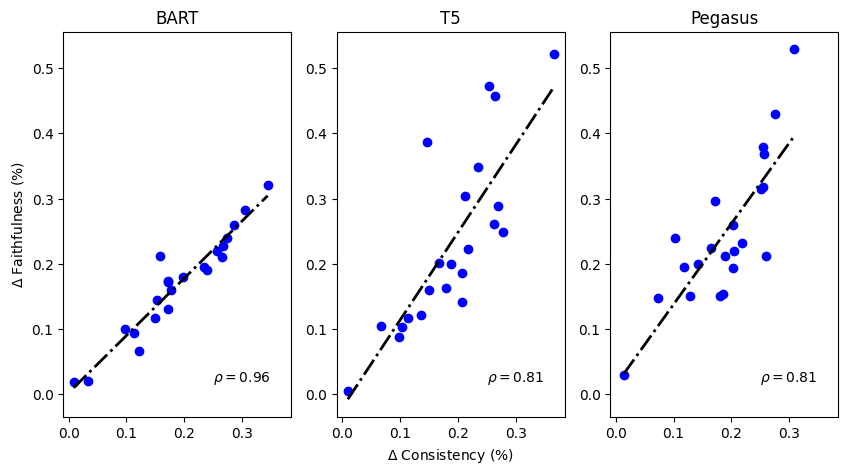}
  \caption{Correlation between faithfulness and consistency dimensions on SAMSum dataset.}
  \label{fig:f_c_sam}
\end{figure}

See Figures~\ref{fig:s_f_sam}, \ref{fig:f_c_sam}, \ref{fig:s_c_sam}.

\subsubsection{TODSum}
\begin{figure}[t]
\centering
\includegraphics[scale=0.3]{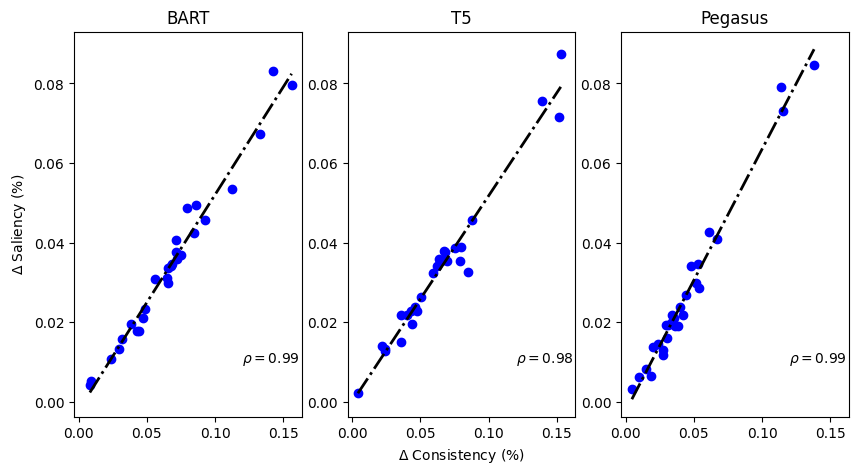}
  \caption{Correlation between consistency and saliency dimensions on TODSum dataset.}
  \label{fig:s_c_tod}
\end{figure}

\begin{figure}[t]
\centering
\includegraphics[scale=0.3]{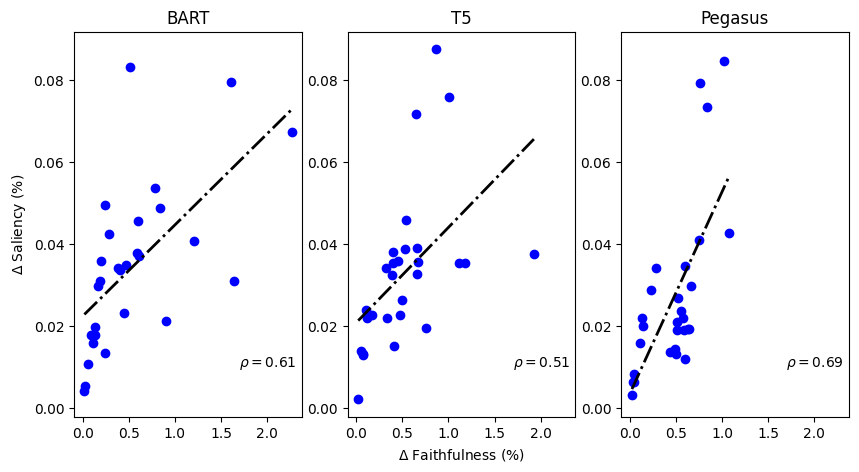}
  \caption{Correlation between faithfulness and saliency dimensions on TODSum dataset (Outliers excluded for the purpose of visualization).}
  \label{fig:s_f_tod}
\end{figure}

\begin{figure}[t]
\centering
\includegraphics[scale=0.3]{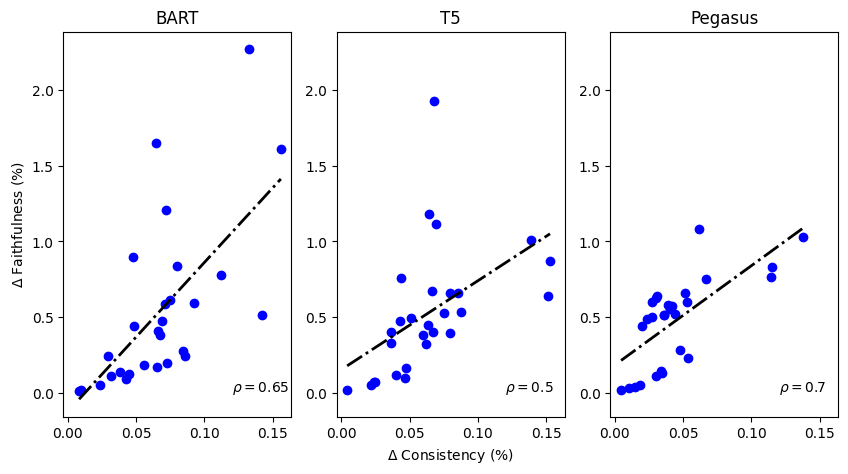}
  \caption{Correlation between faithfulness and consistency dimensions on TODSum dataset.}
  \label{fig:f_c_tod}
\end{figure}

See Figures~\ref{fig:s_f_tod}, \ref{fig:f_c_tod}, \ref{fig:s_c_tod}.

\subsection{Analysis using ROUGE-L and SummaC scores}
\label{subsec:rouge_summac}
\begin{table*}[]
\centering \small
\begin{tabular}{lllllll}
\hline
\multirow{2}{*}{\textbf{Model}} & \multicolumn{3}{c}{\textbf{Utterance Perturbations}}             & \multicolumn{3}{c}{\textbf{Dialogue Perturbations}}              \\ \cline{2-7} 
                                & \textbf{Consistency} & \textbf{Saliency} & \textbf{Faithfulness} & \textbf{Consistency} & \textbf{Saliency} & \textbf{Faithfulness} \\ \hline
BART Large & 14.00$\pm$0.22 & 10.91$\pm$0.01 & 9.18$\pm$0.01  & 14.37$\pm$0.37 & 10.37$\pm$0.01 & 8.97$\pm$0.01  \\
BART Base  & 14.18$\pm$0.29 & 10.65$\pm$0.01 & 9.60$\pm$0.01  & 15.40$\pm$0.31 & 9.74$\pm$0.01  & 9.04$\pm$0.09  \\
Pegasus    & 13.50$\pm$0.46 & 13.24$\pm$0.01 & 11.29$\pm$0.02 & 14.78$\pm$0.39 & 12.14$\pm$0.02 & 9.80$\pm$0.01  \\
T5 Base    & 14.72$\pm$0.36 & 13.43$\pm$0.01 & 11.01$\pm$0.01 & 13.88$\pm$0.42 & 12.27$\pm$0.02 & 9.79$\pm$0.01  \\
T5 Small   & 14.66$\pm$0.33 & 14.40$\pm$0.01 & 10.11$\pm$0.01 & 15.75$\pm$0.31 & 10.99$\pm$0.01 & 8.72$\pm$0.08  \\
DIAL-BART0 & 29.72$\pm$0.36 & 22.70$\pm$0.01 & 20.53$\pm$0.01 & 34.09$\pm$0.30 & 26.3$\pm$0.02  & 23.29$\pm$0.01 \\
FLAN-T5    & 34.06$\pm$0.55 & 34.63$\pm$0.01 & 36.67$\pm$0.02 & 39.84$\pm$0.53 & 36.98$\pm$0.03 & 40.82$\pm$0.06 \\ 
LLAMA-2	& 47.1$\pm$0.17	& 35.16$\pm$0.01 & 33.19$\pm$0.09 & 54.53$\pm$0.48 & 33.59$\pm$0.03 & 31.69$\pm$0.02 \\ \hline
\end{tabular}
\caption{Results on TweetSum using ROUGE-L}
\end{table*}

\begin{table*}[]
\centering \small
\begin{tabular}{lllllll}
\hline
\multirow{2}{*}{\textbf{Model}} & \multicolumn{3}{c}{\textbf{Utterance Perturbations}}             & \multicolumn{3}{c}{\textbf{Dialogue Perturbations}}              \\ \cline{2-7} 
                                & \textbf{Consistency} & \textbf{Saliency} & \textbf{Faithfulness} & \textbf{Consistency} & \textbf{Saliency} & \textbf{Faithfulness} \\ \hline
BART Large & 19.18$\pm$0.35 & 6.66$\pm$0.01  & 3.37$\pm$0.01 & 20.85$\pm$0.60 & 7.70$\pm$0.02  & 2.11$\pm$0.01 \\
BART Base  & 19.35$\pm$0.41 & 6.67$\pm$0.01  & 4.23$\pm$0.02 & 21.08$\pm$0.47 & 5.34$\pm$0.02  & 3.07$\pm$0.01 \\
Pegasus    & 19.67$\pm$0.50 & 8.33$\pm$0.02  & 3.75$\pm$0.01 & 21.70$\pm$0.53 & 7.43$\pm$0.03  & 3.67$\pm$0.03 \\
T5 Base    & 19.20$\pm$0.50 & 7.81$\pm$0.03  & 3.87$\pm$0.03 & 21.40$\pm$0.58 & 7.76$\pm$0.04  & 3.44$\pm$0.01 \\
T5 Small   & 20.77$\pm$0.55 & 8.44$\pm$0.06  & 3.69$\pm$0.01 & 21.17$\pm$0.63 & 5.93$\pm$0.01  & 2.38$\pm$0.04 \\
DIAL-BART0 & 43.05$\pm$0.52 & 12.8$\pm$0.03  & 4.55$\pm$0.01 & 51.75$\pm$0.47 & 16.05$\pm$0.02 & 6.32$\pm$0.03 \\
FLAN-T5    & 39.54$\pm$0.64 & 14.96$\pm$0.00 & 5.95$\pm$0.01 & 45.93$\pm$0.65 & 15.35$\pm$0.04 & 7.72$\pm$0.02 \\ 
LLAMA-2 & 45.05$\pm$0.44 &20.51$\pm$0.04 & 18.06$\pm$0.02 & 56.32$\pm$0.43	& 20.58$\pm$0.11 & 12.79$\pm$0.06 \\ \hline
\end{tabular}
\caption{Results on TweetSum using SummaC}
\end{table*}

\begin{table*}[]
\small \centering
\begin{tabular}{lllllll}
\hline
\textbf{Dimension} &
  \multicolumn{1}{c}{\textbf{Repetitions}} &
  \multicolumn{1}{c}{\textbf{Time Delays}} &
  \multicolumn{1}{c}{\textbf{Greetings}} &
  \multicolumn{1}{c}{\textbf{Conclusion}} &
  \multicolumn{1}{c}{\textbf{Split Utterances}} &
  \multicolumn{1}{c}{\textbf{Combine Utterances}} \\ \hline
Consistency  & 31.03$\pm$0.52 & 25.73 $\pm$0.77 & 36.89$\pm$1.07 & 18.17$\pm$0.95 & 13.34$\pm$0.75 & 8.7$\pm$0.62  \\
Saliency     & 12.16$\pm$0.66 & 9.64$\pm$0.97   & 16.72$\pm$2.36 & 5.62$\pm$0.73  & 11.63$\pm$1.05 & 6.62$\pm$0.77 \\
Faithfulness & 10.17$\pm$0.45 & 7.54$\pm$0.58   & 10.84$\pm$0.93 & 5.3$\pm$0.69   & 8.96$\pm$0.6   & 5.33$\pm$0.49 \\ \hline
\end{tabular}
\caption{Impact of Dialouge Perturbations on TweetSum using ROUGE-L}
\end{table*}

\end{document}